\documentclass[letterpaper, 10 pt, conference]{ieeeconf}

\IEEEoverridecommandlockouts

\usepackage{lipsum}
\usepackage{graphicx}
\usepackage{amssymb}
\usepackage{amsmath}
\usepackage{tcolorbox}
\usepackage{algorithm}
\usepackage{algpseudocode}
\usepackage{cite}
\usepackage{bbm}
\usepackage{balance}

\usepackage[english]{babel}
\usepackage[autostyle, english = american]{csquotes}

\usepackage{etextools}
\usepackage{amssymb}
\usepackage{bm}
\usepackage{float}
\usepackage{makeidx}
\usepackage{amsmath}
\usepackage{bbm}
\usepackage{epsfig}
\usepackage{epsf}
\usepackage{psfrag}
\usepackage{verbatim}
\let\labelindent\relax
\usepackage{enumitem}
\usepackage{color}
\usepackage{multirow}
\usepackage{tabularx}
\usepackage{booktabs}
\usepackage[tight,footnotesize]{subfigure}
\usepackage{array}
\usepackage{soul}
\usepackage{footnote}
\usepackage{cite}
\usepackage{dblfloatfix}

\usepackage{color, colortbl}
\usepackage{graphicx}
\graphicspath{{Figures}}
\DeclareGraphicsExtensions{.pdf,.png}
\usepackage{bigstrut}
\usepackage[english]{babel}

\usepackage{csquotes}

\usepackage[T1]{fontenc}
\usepackage{algorithm, setspace}
\usepackage{algpseudocode}
\usepackage{url}
\usepackage{multirow}
\usepackage{float}
\usepackage{xcolor}
\usepackage{hyperref}
 \hypersetup{
     colorlinks=true,
     linkcolor=orange,
     filecolor=orange,
     citecolor=orange,      
     urlcolor=orange,
     }

\usepackage{balance}

\MakeOuterQuote{"}

\newcommand{\p}[1]{\medskip \noindent \textbf{{#1}.}}
\newcommand{\eq}[1]{Equation~(\ref{eq:#1})}
\newcommand{\fig}[1]{Figure~\ref{fig:#1}}

\makeatletter
\newcounter{phase}[algorithm]
\newlength{\phaserulewidth}

\makeatother

\title{\LARGE 
Fine-Tuning Robot Policies While Maintaining User Privacy
}

\author{Benjamin A. Christie, Sagar Parekh, and Dylan P. Losey
\thanks{This work is supported in part by NSF Grant $\#2246446$. The authors are members of the Collaborative Robotics Lab (\href{https://collab.me.vt.edu/}{Collab}), Dept. of Mechanical Engineering, Virginia Tech, Blacksburg, VA 24061. \newline Corresponding author's email: \texttt{benc00@vt.edu}}
}
\begin{document}
\maketitle
%%%%%%%%%%%%%%%%%%%%%%%%%%%%%%%%%%%%%%%%%%%%%%%%%%%%%%%%%%%%%%%%%%%%%%%%%%%%%%%%

\begin{abstract}

Recent works introduce general-purpose robot policies.
These policies provide a strong prior over how robots should behave --- e.g., how a robot arm should manipulate food items.
But in order for robots to match an individual person's needs, users typically \textit{fine-tune} these generalized policies --- e.g., showing the robot arm how to make their own preferred dinners.
Importantly, during the process of personalizing robots, end-users leak data about their preferences, habits, and styles (e.g., the foods they prefer to eat).
Other agents can simply roll-out the fine-tuned policy and see these personally-trained behaviors.
This leads to a fundamental challenge: how can we develop robots that \textit{personalize} actions while keeping learning \textit{private} from external agents?
We here explore this emerging topic in human-robot interaction and develop \textit{PRoP}, a model-agnostic framework for personalized and private robot policies.
Our core idea is to equip each user with a unique key; this key is then used to mathematically transform the weights of the robot's network.
With the correct key, the robot's policy switches to match that user's preferences --- but with incorrect keys, the robot reverts to its baseline behaviors.
We show the general applicability of our method across multiple model types in imitation learning, reinforcement learning, and classification tasks.
PRoP is practically advantageous because it retains the architecture and behaviors of the original policy, and experimentally outperforms existing encoder-based approaches.
%See videos and code here: \url{https://prop-icra26.github.io}

\end{abstract}

%%%%%%%%%%%%%%%%%%%%%%%%%%%%%%%%%%%%%%%%%%%%%%%%%%%%%%%%%%%%%%%%%%%%%%%%%%%%%%%%

\section{Introduction} \label{sec:intro}

Generalist policies enable robots to learn multiple tasks \cite{black2024pi_0, li2024cogact}.
So far these methods have traditionally been used in research labs and factories.
But we envision a future where robots enter domestic settings for assisting humans \cite{kawaharazuka2024real}.
For example, consider a robot that is developed to help in a kitchen.
This robot will have some initial policy $\pi_0$ that users may want to finetune to match their own preferences and requirements.
For instance, perhaps the robot knows how to make a hamburger, but individual users prefer different ingredients, condiments, or even specific sanitation procedures.
This finetuning raises privacy concerns: the manufacturers can share the users' data collected during finetuning with third-parties.
Consequently, there is increasing demand for exploring new avenues to maintain the privacy and transparency of robotic agents \cite{news}.
Following this, we come to a fundamental scientific question: how do we make systems that can learn and adapt to individual end-users, while still maintaining those user's privacy?

\begin{figure}
    \centering
    % \vspace{-1.5em}
    \includegraphics[width=1\linewidth]{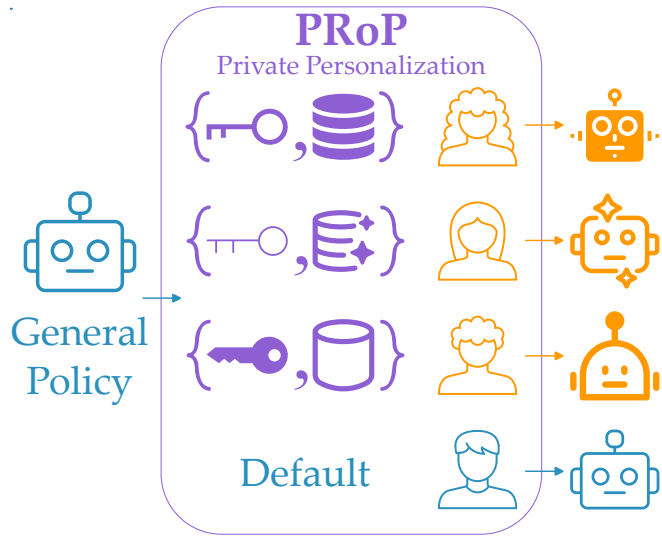}
    %\caption{(Top) In human-robot interaction, robots are often finetuned to personalize to user-specific needs. The two users shown in above have different preferences for how they like their burger. The robot, which is trained on a generalized recipe, can be personalized to different users. However, this personalization can be extracted from the robot's policy by mimicking the user actions and observing the policy outputs. This is a major privacy concern.
    %(Bottom) In this work, we present a method for user personalization while ensuring privacy at the interaction level though the use of keys such as voice recognition. Under our framework, unauthorized people cannot mimic the user-specific keys. This ensures that the privacy of users is secured. Notably, our method achieves this without altering the original network architecture.}
    \caption{In human-robot interaction robots are often finetuned to personalize to user-specific needs. 
    The users above have different preferences encoded in their personalized datasets. 
    When the general model is finetuned to the user's personalized dataset, the resulting policy is not \textit{private}. Any user that interacts with the finetuned policy will be able to infer the user's preferences.
    Instead, we propose \textbf{PRoP}: a method that enables {\textbf{p}rivate personalization of \textbf{ro}bot \textbf{p}olicies to humans}. 
    PRoP learns to associate user keys with intermediate transformations of the original policy, causing personalized and private behavior.
    When users do not provide a key (or provide a key not included in PRoP's training set), they receive the original policy.
    %Given a pretrained policy, PRoP learns to associate user keys with intermediate transformations of the original policy, causing personalized output.
    }
    \label{fig:front}
    \vspace{-1.5em}
\end{figure}

Privacy in machine learning has traditionally been examined from two perspectives.
First is data privacy, which concerns safeguarding the sensitive information of individuals represented in the dataset \cite{li2019differentially, zhang2021leakage, chen2020comprehensive, abadi2016deep}.
Second is model privacy, which focuses on protecting the learned parameters of a neural network through techniques such as {homomorphic} encryption or differentially-private learning \cite{lytvyn2019information, lee2022privacy, pulido2021privacy, triastcyn2020bayesian}.
In this work, we adopt a third perspective with respect to robot learning: ensuring that a trained, personalized robot does not leak user preference information to other users.
Returning to our example, privacy in this context means that the robot can be finetuned to learn your preferred way of making a recipe while preventing unauthorized users from accessing those preferences \textit{even if they have access to the trained model}.
In practice, this can be difficult to achieve because --- if someone has access to the finetuned model --- they can roll-out this model and infer the previous user's preference by watching the robot actions.
So how do we safeguard privacy of user preferences?
Our insight is that:
\begin{center}
    \textit{We can use latent values to} transform \textit{intermediate features of a network for enhanced privacy and personalization.}
\end{center}

Concretely, we leverage \textit{keys} (\fig{front}).
A key is any feature that is unique to the user such as facial structures, vocal patterns, or a textual password.
When finetuning the robot under {our} approach, a user combines their unique key with the intermediate features of the network and trains it to output their personalized actions.
This unique mechanism for personalizing robots safeguards user privacy since preference information remains inaccessible to anyone who does not have the user's key.
Without careful design, keys may unintentionally cause the robot to forget its general-purpose policy.
But our technical approach avoids this pitfall --- and preserves the initial model architecture --- by leveraging the key to perform mathematical operations on the intermediate weights. 
Our proposed mechanism is not tied to a specific network architecture or application as we later demonstrate in our experiments with visual data, imitation learning, classifiers, and reinforcement learning.
Indeed, as shown in our experiments on robot arms, users can finetune the robot to make their desired recipes without losing the robot's previously learned behaviors, and without exposing their preference to other agents.
We see this work as a step towards safe and personalized human-robot interaction.

Overall, we make the following contributions: 

\p{Key-based Personalization of Robot Policies}
We present a formulation for key-based personalization of robot control policies. 
Under this formalism, the robot learns to personalize to new users' specifications while retaining its original, general behavior. 
This formalism is nontrivial to implement in a learning algorithm, since the original and conditional policies operate in different domains, i.e., adding a key as the input requires changing the size of the pre-trained architecture.
Instead, we use keys to transform the intermediate features of the pre-trained policy, circumventing the need for changing the architecture size.

\p{Personalized and Private Robot Policies}
We present our implementation of the aforementioned key-based personalization with privacy guarantees. 
Our method, \textbf{PRoP} (Personalized and Private Robot Policies) retains the original network architecture, exhibits behavior of the original robot policy for unprivileged users, and personalizes to specific users through a privacy-oriented mechanism.
Importantly, {PRoP} extends to arbitrary learning rules and architectures that enables simple, end-to-end training of the model. 

\p{Real-world Validation and Empirically Verified Robustness}
We empirically test the performance of {PRoP} in a collection of controlled simulations and real-world studies, including Imitation Learning, Reinforcement Learning, Image Classification, and Task Allocation. 
We further extend {PRoP} to more complex settings, such as language prose personalization and key-based obfuscation. 
\section{Related Works}

In recent years there has been a variety of research that analyzes robot-human personalization. 
This is especially true in Human-Robot Interaction (HRI): humans can provide additional information to robots in order to personalize future interactions. 
This information may be demonstrations \cite{jevtic2018personalized, christie2024limit}, corrections \cite{losey2022physical}, preferences \cite{gasteiger2023factors}, or any combination of the three \cite{mehta2024unified, jain2015learning}.

Within these settings the robot adjusts its parameters to align with the user's desired behaviors. 
However, most existing research on human-to-robot personalization ignore the dimension of privacy. 
While some works exist that outline approaches to protecting user privacy in HRI \cite{chatzimichali2020toward,lee2011understanding}, these primarily focus on protecting the data gathered from users.
But even if the robot's training set is secure, once a robot's policy has been personalized, any human with access to the policy will be able to infer the previous operator's preferences.
For this reason we focus on privacy not at the data level --- which has been explored --- but at the \textit{interaction} level, ensuring that user information remains protected during human-robot interaction.

\p{Personalizing Policies}
Robots often learn policies instantiated as neural networks.
Once trained on a general dataset, neural networks can then be finetuned (i.e., personalized) to new distributions through data aggregation. 
As previously mentioned, methods can use diverse data types to personalize to new users: learning from preferences, corrections, and demonstrations is standard practice \cite{jevtic2018personalized, christie2024limit, losey2022physical, gasteiger2023factors, jain2015learning}.
The personalized network is learned end-to-end and will affect \textit{all} future users: it is not gated to that particular user.
Put simply, subsequent users that have the fine-tuned model can potentially infer the preferences of previous users by directly interacting with the robot.

A similar challenge exists in fine-tuning pretrained policies for novel tasks.
Common approaches for fine-tuning involve using skill extraction and transfer \cite{kim2024robust,pmlr-v164-hundt22a}, optimization of learned reward models \cite{10610421,li2024process,yu2024b}, or strategically collecting samples to maximize information gain \cite{bagatella2024active,hubotter2024active,julian2020never}.
While these methods focus on cross-task generalization, they essentially adapt pretrained policies to new user preferences encountered during deployment.
Importantly, these policies are not explicitly designed to obfuscate or decouple learned behaviors.
This transparency introduces significant privacy risks, as a malicious actor can potentially access private preferences of previous users.
To provide privacy guarantees, it is necessary to implement a gating mechanism that restricts access to learned behaviors based on user authorization.
To this end, we propose a method that safeguards user privacy by associating each user with a unique latent key, restricting their access to their respective behaviors.
We recognize that standard model architectures are fixed and cannot be easily transferred to a new domain of inputs. For example, a policy trained on states as inputs cannot be easily restructured to accept additional inputs, i.e., authorization key, without significant retraining.
Therefore, our method avoids architectural changes and instead transforms the intermediate feature representations of the network.
This makes our method highly compatible with pretrained models, enabling private personalization while retaining the original learned behaviors.

\p{Human-Robot Privacy}
Existing works on user privacy in HRI typically deal with securing the user's underlying training data: whether that be their demonstrations or demographic information. 
\cite{chatzimichali2020toward} presents methods for securing user datasets for robot learning, and \cite{lee2011understanding} addresses how people \textit{perceive} the way their data is utilized in HRI research. 
On a lower level, works such as \cite{abadi2016deep, li2019differentially, triastcyn2020bayesian} employ $(\epsilon, \delta)$-differentially private learning to assert privacy guarantees on user-data.
Unfortunately, approaches that use differentially-private learning still struggle with budgeting $\epsilon$ for complex interaction tasks. 
For example, the state-of-the-art presented in \cite{triastcyn2020bayesian} still exhibits a probability of privacy-failure of approximately $67.2\%$ on the MNIST classification task. 
Furthermore, these methods are not gated: they attempt to secure \textit{training data} instead of securing the policy's \textit{output}.

Methods that leverage fully homomorphically-encrypted neural networks \cite{lee2022privacy, lytvyn2019information} may solve our problem, but they are computationally infeasible. 
From our testing, fully-homomorphically encrypted inference can be up to $10^7$ times slower than baseline, which is infeasible for real-time robotics applications.  
Instead, we propose that methods that \textit{mimic} encryption by obfuscating the intermediate features of the policy network can be used for private human-robot personalization.
Overall, our proposed method will lock the user's preferences behind their key, and even trying mathematically similar keys will not cause the network to output the user's fine-tuned behaviors. 

\section{Problem Statement}\label{sec:problem}

We are interested in human-robot interaction settings where a pre-trained robot policy --- represented as a neural network --- must be personalized to specific users and remain unchanged for others. 
In this manuscript, we interchange ``user'' and ``human collaborator,'' but in practice the ``user'' could be another robotic agent.

\p{Dynamics}
The robot takes actions $u \in \mathcal{U}$ according to its policy $\pi$:
\begin{equation}
u \sim \pi\left(\circ \mid x\right)
\end{equation}
where $x \in \mathcal{X}$ is the system state. 
The system state transitions in discrete-time with deterministic dynamics according to the state-transition function $f: \mathcal{X}\times\mathcal{U} \mapsto \mathcal{X}$:
\begin{equation}
    x^{t+1} = f(x^t, u^t) = f\left(x^t, 
    \{u_{i}^t\}_{i = 0}^N
    \right)
    \label{eq:transition}
\end{equation}
Here $t \in [0, T)$ represents the current timestep. 
More generally, the system may transition according to the actions of multiple agents. 
In this general case $u$ in \eq{transition} becomes the combined actions of $N$ robots and $M$ humans, i.e., $u = u_{R_1} \cup u_{R_2} \cdots \cup u_{R_N} \cup u_{H_1} \cdots \cup u_{H_M}$, where $u_{R_i}$ is the action of robot $i$ and $u_{H_j}$ is the action of human $j$.
%In environments with a single robot and human, $u^t = \{u^t_\mathcal{H}, u^t_\mathcal{R}\}$ where $u^t_\mathcal{H}$ is the human's action at timestep $t$. 

\p{Personalization of Pretrained Policies}
We assume that pretrained policies are represented as neural networks.
These neural networks are parameterized by weights $\theta$ which follow the gradient descent learning rule shown below:
\begin{equation}
    \theta^{\tau + 1} = \theta^\tau - \alpha \nabla_{\theta} \mathcal{L}(\theta^\tau)
    \label{eq:back}
\end{equation}
where $\mathcal{L}$ is the loss function to be minimized by $\theta$.
Traditionally, personalization or finetuning of existing policies occurs by retraining the policy on a new dataset or loss function \cite{mehta2024unified, hu2023llm, yang2024robot, christie2024limit}. 
For example, on a user-specific level, finetuning the original weights $\theta^0$ can cause the general model to better fit the user's preferences \cite{mehta2024unified}, objectives \cite{hu2023llm, yang2024robot}, or perceptions \cite{christie2024limit}. 
Hereafter we refer to this pretrained robot policy with weights $\theta^0$ as $\pi^\star$.

\p{Private Personalization}
Above we outlined a simple dynamical system where the robot's action $u$ is sampled from a distribution conditioned on the system state $x$. 
However, for the robot to personalize to multiple users, the robot policy should instead condition on the system state \textit{and} some personal user information. 
Hence, the personalized policy is a distribution conditioned on the state and user information:
\begin{equation}
    u \sim \pi(\circ \mid x, k)
    \label{eq:cond-policy}
\end{equation}
where $k \in \mathcal{K}$ is the personal information that the robot should condition their behavior on. 
This personal information is assumed to be non-fungible.
In other words, two separate users should not be able to imitate each other's $k$.
This information can take many forms, such as biographical information, facial features, fingerprints, or a password.
In this work we treat $k$ as the bit-representation of a user password.
Note that representing the policy as a conditional distribution on $k$ is critical from a privacy perspective. 
A naive implementation of a privately personalized policy is to have a separate robot policy $\pi_0(x), \ldots, \pi_N(x)$ for each user $\mathcal{H}_0, \ldots, \mathcal{H}_N$.
If the developer has access to policy $\pi_i$, then they can directly infer the private preferences of the user $\mathcal{H}_i$: this becomes a clear breach of user privacy.

We seek to learn a robot policy that follows the form of \eq{cond-policy} while leveraging the architecture and weights of the pretrained policy $\pi^\star$. 
Satisfying both of these requirements is non-trivial: we assume that the pretrained policy has a mapping $\pi^\star: \mathcal{X} \mapsto \mathcal{U}$ while the personalized robot policy has a mapping $\pi^{p}: \mathcal{X} \times \mathcal{K} \mapsto \mathcal{U}$. 
Furthermore, the amended domain should not interfere with the model performance. 
For example, performance of the personalized model with an incorrect key should be as close as possible to the general policy $\pi^\star$.
In what follows, we will discuss our method for privately personalizing a pretrained policy without affecting its architecture or base behavior.

\section{Personalized and Private Robot Policies}\label{sec:method}

\begin{figure}
    \centering
    \includegraphics[width=1.0\linewidth]{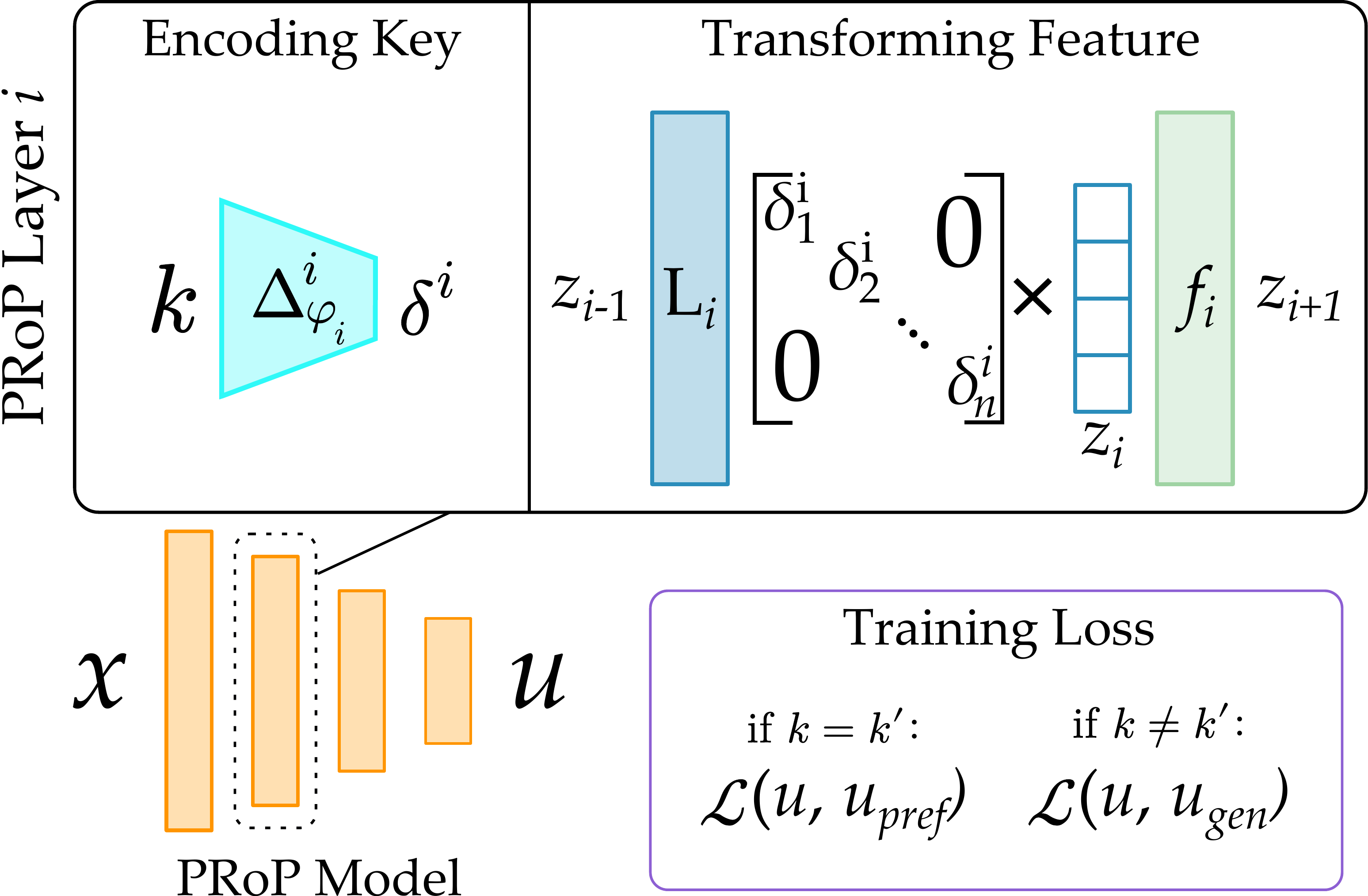}
    \caption{Schematic diagram of PRoP. Our method for private personalization of robot policies uses a key encoder to augment the intermediate features of the neural network $\mathcal{R}_\phi$. 
    Particularly, at the intermediate layer $i$ of the original policy $\pi^\star$, we apply an affine transformation to the features $z_i$ using \eq{augment}. This transformation is shown in the top row.
    It is noted that this augmentation does not need to occur at every interstitial layer of the neural network: we find in our controlled simulations that a single application of the PRoP mechanism is sufficient for personalization. 
    When combined with a conditional, personalized loss function (shown in bottom right), we find that PRoP outperforms baseline algorithms in terms of \textit{privacy} and \textit{personalization} without changing the architecture of the original policy $\pi^\star$.
    }
    \vspace{-2.0em}
    \label{fig:method}
\end{figure}

In this section we introduce \textbf{PRoP}: \textbf{P}rivate \textbf{Ro}bot \textbf{P}olicies.
PRoP implements \eq{cond-policy}, integrating user keys into the original robot policy $\pi^\star$ without affecting its architecture.
Towards this end, we leverage encoders that map the human's keys into a latent space, and then use the latent encodings to transform the intermediate features of the robot policy architecture. 
Our method is summarized in \fig{method}.

Concretely, we train the key encoders $\Delta^i_{\varphi_i}: \mathcal{K} \mapsto \mathcal{Z}^i \in \Delta^{0:N}_{\varphi}$ corresponding to each of the intermediate layers of the robot policy network.
The encoders are parameterized by $\varphi_i$, and since the latent encoding transforms the intermediate features of the policy architecture, the size of the latent space $\lvert\mathcal{Z}^i\rvert$ is determined by the pretrained policy's intermediate architecture at layer $i$. 
These encoders are coupled with the robot's policy $\mathcal{R}_\phi: \mathcal{X} \mapsto \mathcal{U}_\mathcal{R}$ to perform private personalization using keys $k \in \mathcal{K}$.
To ensure personalization without affecting the architecture of $\mathcal{R}_\phi$, we perform latent augmentation using an affine transformation at select hidden layers of $\mathcal{R}_\phi$. 
This process will be described in more detail below. 
It should also be noted that since $\mathcal{R}_\phi$ has the same domain as the original policy, it is possible that a third party simply would not use the encoders $\Delta^{0:N}_\varphi$. 
If this is the case, then $\mathcal{R}_\phi$ should mimic the original pretrained policy $\pi^\star$. 
Put another way: if a user does not provide a key, then the robot should behave according to the (general) pretrained policy.

\p{Prerequisites}
We assume access to the original objective $\mathcal{J}^\star$ and loss function $\mathcal{L}^\star$ used to train $\pi^\star$. 
We additionally assume that $\pi^\star$ is parameterized by weights $\theta$ that are updated using backpropagation, like in \eq{back}.
We will modify the loss function $\mathcal{L}^\star$ of \eq{back} for the private personalization of $\pi^\star$ to a new, personalized objective $\mathcal{J}^\prime \ne \mathcal{J}^\star$. 
Returning to our motivating example, objective $\mathcal{J}^\star$ corresponds to the general recipe of the dinner that the robot policy knows, and the personalized objective $\mathcal{J}^\prime$ corresponds to the user's preferred dinner recipe.
Our method for private personalization of the policy $\pi^\star$ should not modify the architecture of $\pi^\star$: we assume that there are linear layers within the original network architecture that are publicly exposed or otherwise copyable: such as in multi-layer perceptrons, transformers, or the feature layers of convolutional neural networks. 
This condition is necessary for our method to apply transformations to the intermediate features of the original network.

\p{Key Encoding}
Each key encoder $\Delta^i_{\varphi_i} \in \Delta^{0:N}_\varphi$ is a multi-layer perceptron with weights $\varphi_i$ and final activation function $\tanh$. 
The key encoder takes in the user-specific key and maps that to a latent value used to personalize the network. The key encoder operates across users and is unique to a specific hidden layer of the policy network $\mathcal{R}_\phi$.
The output size of each key encoder should correspond with the output dimension of select hidden layers of the neural network $\mathcal{R}_\phi$. 
For example, take $\mathcal{R}_\phi$ to be a neural network with two layers, hidden dimension $a$, and nonlinear activation function $f$. In this case, $\lvert \Delta^{0:N}_\varphi \rvert = 1$ and $\lvert \mathcal{Z}_1 \rvert = a$. 
The key encoding performs a intermediate affine transformation of the policy network as follows.
Take $W_i$ and $b_i$ to be the weight and bias of the $i$-th layer of $\mathcal{R}_\phi$ and $\Delta^i_{\varphi^i}(k) = \delta_i$.
The resulting output of the $i$-th layer is:
\begin{equation}
    z_{i + 1} = f\left(W_i~ \text{diag}\left(\delta_i\right) z_i + b_i\right)
    \label{eq:augment}
\end{equation}
{This weight-transformation architecture is intentionally chosen over alternatives such as using the key as a standard input feature or maintaining a discrete library of user-specific weights. By embedding the personalization into a transformation of the base weights $\phi$, PRoP ensures that private behaviors are {entangled} within the network. This provides a layer of weight-level obfuscation: without the correct key, the personalized policy is mathematically inaccessible. Furthermore, this approach allows for sub-linear parameter scaling, as multiple user preferences can be 'compressed' into the shared base weights rather than requiring separate storage for each user.
}
Applying this transformation directly to the policy network's hidden layers modifies the relationship between the weights and output of the policy, enabling us to utilize the architecture of $\pi^\star$ while conditioning the policy on a specific user's key.
Additionally, if we omit $\delta_i$, then the policy $\mathcal{R}_\phi$ will resolve to default behavior: ideally, the pretrained policy $\pi^\star$.
When rolling out $\mathcal{R}_\phi$, we will perform \eq{augment} for specific hidden layers. We annotate this process as $\mathcal{R}_{\phi\cup\varphi}: \mathcal{X} \times \mathcal{K} \mapsto \mathcal{U}_\mathcal{R}$. 
%Note that --- as previously stated --- an agent may run the PRoP policy without using a key. 
%To unify this, we assume that the space $\mathcal{K}$ contains a null element $\varnothing$. 
%When the null element is used as an input to PRoP, $\text{diag}\left(\delta_i\right) = I$ by convention since $W_i z_i = W_i I z_i$ trivially holds. 
%This ensures that if the policy is used without a key, it will follow the default behavior of $\pi^\star$.
This transformation is visually annotated in \fig{method}.

\p{Policy}
Now that we have a structure for PRoP, our next step is to update the weights $\phi$ and $\varphi$ to facilitate private personalization.
Before augmenting PRoP's policy, we ensure that $\mathcal{R}_\phi$ has the same weights as the pretrained policy $\pi^\star$ (i.e., we ensure that $\phi = \theta$). 
Recall that we have access to the original loss function $\mathcal{L}^\star$ and objective $\mathcal{J}^\star$. 
Here we introduce an auxiliary loss function $\mathcal{L}^\prime$ that is identical to $\mathcal{L}^\star$, except that it operates on $\mathcal{R}_{\phi\cup\varphi}$ using the latent multiplication shown in \eq{augment} for a particular key $k$.

Now we begin training PRoP. 
Given a user with key $k^\prime$ and preferences encoded in their objective $\mathcal{J}^\prime$, we can update the policy and key encoders using the following loss function: 
\begin{equation}
\mathcal{L}\left(\phi, \varphi\right)
=
\mathcal{L}^\prime_{k^\prime, \mathcal{J}^\prime}\left(\phi, \varphi\right)
+
\sum\limits_{k \in \mathcal{K} - \{k^\prime\}}
\mathcal{L}^\prime_{k, \mathcal{J}^\star}\left(\phi, \varphi\right)
\label{eq:prop}
\end{equation}
Note that $\varnothing \in \mathcal{K}$, so the resulting policy $\mathcal{R}_{\phi\cup\varphi}$ should mimic $\pi^\star$ when the key is omitted from \eq{augment}.
The loss function in \eq{prop} assumes that there is a single key that should yield personalized behavior; operating on a set of personalized behaviors is not very different.
Assuming a collection of users each with keys $k_i^\prime \in \mathcal{K}^\prime \subset \mathcal{K}$ and objectives $\mathcal{J}^\prime_i \ne \mathcal{J}^\star$, we use the following loss function:
\begin{equation}
\mathcal{L}\left(\phi, \varphi\right)
=
\sum\limits_{k^\prime \in \mathcal{K}^\prime}
\mathcal{L}^\prime_{k^\prime, \mathcal{J}_i^\prime}\left(\phi, \varphi\right)
+
\sum\limits_{k \in \mathcal{K} - \mathcal{K}^\prime}
\mathcal{L}^\prime_{k, \mathcal{J}^\star}\left(\phi, \varphi\right)
\label{eq:prop2}
\end{equation}
In practice, enumerating over $\mathcal{K} - \mathcal{K}^\prime$ in Equations~(\ref{eq:prop}) and (\ref{eq:prop2}) is intractable due to the size of $\mathcal{K}$.
For an extreme example: if $\mathcal{K}$ represents the space of $8$ character passwords and evaluating $\mathcal{L}^\prime$ takes $1$~ns, a single epoch of \eq{prop2} would take more than $500$ years (assuming single-threaded execution). 
To tractably approximate \eq{prop2}, we use an inductive loss function over specific subsets of $\mathcal{K}$:
\begin{equation}
    \begin{gathered}
        \mathcal{K}_1 = \{k \mid k \in \mathcal{K} - \mathcal{K}^\prime, \exists k^\prime \in \mathcal{K}^\prime  ~\text{s.t.}~ \| k - k^\prime\| \le \epsilon \}
        \\
        \mathcal{K}_2 = \{k_i \mid k_i \sim U\left[\mathcal{K} - \mathcal{K}^\prime\right], i \in [1, N_k]\}
        + \{\varnothing\}
        \\
        \mathcal{L}\left(\phi, \varphi\right)
        =
        \sum\limits_{k^\prime \in \mathcal{K}^\prime}
        \mathcal{L}^\prime_{k^\prime, \mathcal{J}_i^\prime}\left(\phi, \varphi\right)
        +
        \sum\limits_{k \in K_1 \cup K_2}
        \mathcal{L}^\prime_{k, \mathcal{J}^\star}\left(\phi, \varphi\right)
        \label{eq:prop3}
    \end{gathered}
\end{equation}
where $\epsilon$ and $N_k$ are hyperparameters chosen by the designer. 
{The subset $K_2$ provides a stochastic approximation of the `global` key space, ensuring that the model converges towards the fallback policy for unauthenticated users, similar to ideas found in negative sampling \cite{yang2024does}. However, unlike standard negative sampling which seeks an unbiased estimate, our inclusion of the subset $K_1$ acts as a form of importance sampling that reduces the variant of the gradient at the boundary of $k \in K^\prime$. This forces the model to learn a high-margin separation between authorized and unauthorized access --- a requirement for privacy that uniform sampling alone would fail to capture.
}
%\eq{prop3} resembles an inductive loss over $\mathcal{L}^\prime$ and approaches the intractable loss function shown in \eq{prop2}. 
%as both $\epsilon \to \infty$ and $N_k \to \infty$. 

\p{Implementation Details}
Above we state that PRoP uses a collection of encoders to privately personalize the robot policy to different users. However, as we will empirically show, using a single key encoder and applying a single intermediate transformation for a small-sized robot policy is sufficient. 
Additionally, as we will show in Section~\ref{sec:sims}, PRoP can be used without a pretrained policy $\pi^\star$: the weights $\phi$ and $\varphi$ can be learned simultaneously in an end-to-end manner. 
{Note that in this work we use $\mathcal{K} = \{0, 1\}^N$, but PRoP is not restricted to this key space.}
A sample implementation of PRoP is available \href{https://prop-icra26.github.io/}{here}.

\section{Experimental Validation}\label{sec:sims}

\begin{figure}
    \centering
    \includegraphics[width=1.0\linewidth]{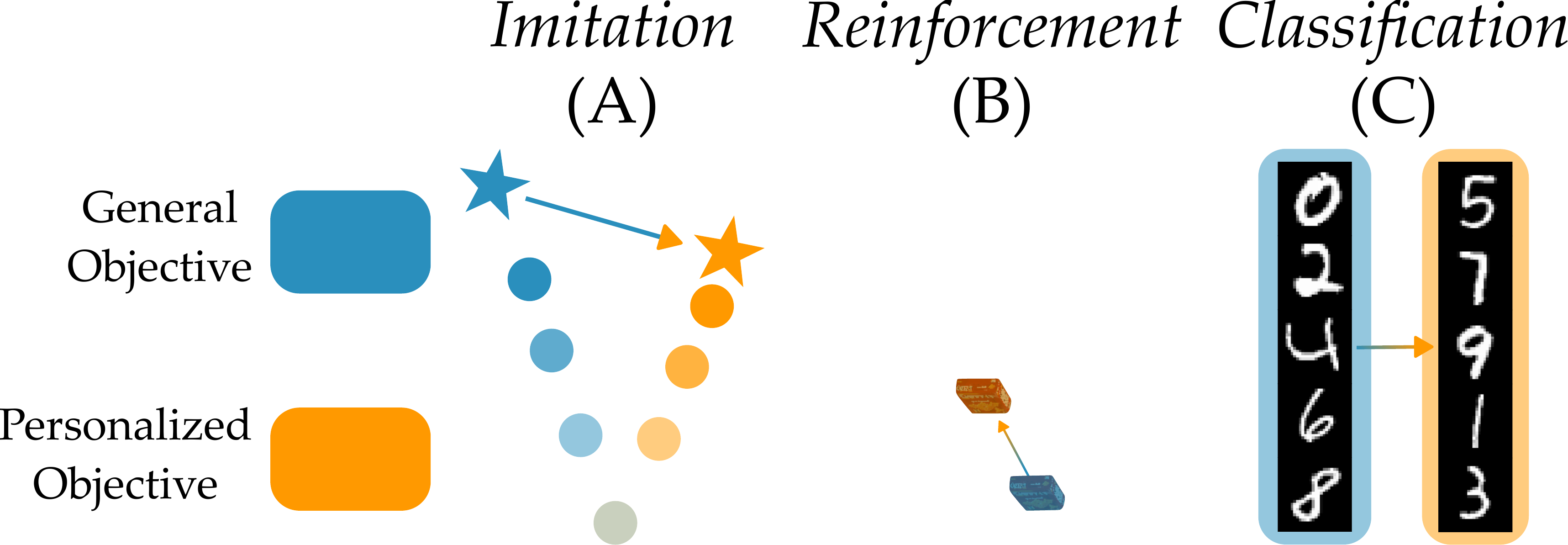}
    \caption{A depiction of our simulated environments presented in Section~\ref{sec:sims}. (A) In \textit{Imitation Learning}, the robot should learn to take actions in the general and personalized datasets, depending on the key. (B) In \textit{Reinforcement Learning}, the robot should learn to move to reach different objects in the scene depending on the key. (C) In \textit{Image Classification}, the policy learns different labels depending on the key.}
    \label{fig:envs}
    \vspace{-1.0em}
\end{figure}

To assess how our method performs compared to existing literature, we conducted an ensemble of experiments across simulated environments. 
The baselines we chose for these experiments are standard architectures commonly seen in human-robot interaction.
The simplest baseline (\textbf{MLP}) is an end-to-end multi-layer perceptron that has an input dimension of $\lvert \mathcal{X} \rvert + \lvert \mathcal{K} \rvert$. 
The hidden dimension of the {MLP} is chosen such that the number of trainable parameters is as close as possible to the number of parameters that our method uses. 
The second baseline (\textbf{CVAE}) is the conditional-variational autoencoder presented in \cite{sohn2015learning}. 
In preliminary testing we implemented two variants: conditioning on the state $x$ and conditioning on the key $k$. 
We found that conditioning on the state was far more performant; we present those results in this section. 
Finally, we compare these baselines to our approach for personalizing to humans presented in Section~\ref{sec:method} (\textbf{PRoP}). 
The environments used are shown in \fig{envs}.

\subsection{Imitation Learning}\label{sec:bc}
In this experiment we prepare a dataset of $N$ expert demonstrations $\mathcal{D} = \{(x, u)^0, \ldots, (x, u)^N\}$. 
The state $x \in \mathbb{R}^{2 n}$ is the robot position concatenated with a goal position $g \in \mathbb{R}^n$. 
The environment follows linear state feedback dynamics. 
The \textit{personalized} objective has a modified set of expert demonstrations that navigate to a different position in the environment parameterized by $g$.
Instead of the expert robot moving to $g$, it moves to a position in the environment offset $g$ by an arbitrary but fixed affine transform. 
For these environments we chose a key size of $128$ (i.e., $\mathcal{K} = \{0, 1\}^{128}$). 
To ensure that the additional weights $\varphi$ do not give {PRoP} an advantage, each model architecture is standardized such that the total parameter count is about $50$k.

\subsection{Reinforcement Learning}
In the second simulated environment we conducted reinforcement learning simulations in the \texttt{PandaGym} environment \cite{gallouedec2021pandagym} using a PPO-style Actor-Critic architecture and loss function \cite{schulman2017proximal} with joint-space control. 
To improve convergence of all methods, {we changed} the reward function for the \textit{Reach} environment {to a dense variant}:
\begin{equation}
    R(x, u) \propto 
    \left\| x_\mathcal{R}^\text{ee} - g\right\| - \| f(x, u)_\mathcal{R}^\text{ee} - g \|
\end{equation}
We have found that if the reward function is not normalized, PPO-style critic algorithms fail to converge over receding horizons regardless of architecture. 
Similar to the \textit{Imitation Learning} environment, we modify the location of $g$ according to an arbitrary but fixed affine transformation that is constant for a particular experiment. 
We normalize the network architecture to have approximately $100$k weights for each method and use a key size of $128$.

\subsection{Image Classification}\label{sec:mnist}
To assess how our method extends to settings without a state-transition, we conducted a classification test on the MNIST dataset \cite{deng2012mnist}. As is standard in image classification, the default (i.e., unpersonalized) behavior is to minimize the cross-entropy loss of the predicted label with the ground-truth label. The personalized behavior is to predict the ground-truth label $l$ subject to an offset according to $\left(l + k\right) \mod 10$ where $k \in \mathbb{Z}^+ $ is an arbitrary but fixed positive integer. 
% As in previous experiments, we standardize each model architecture to an approximate parameter count of $100$k. 
Each model architecture is a sequence of convolutional layers with batch normalization, followed by a sequence of linear layers for feature extraction.

{
\subsection{Personalization Capacity}\label{sec:cap}
The experiments in Sections~\ref{sec:bc} through \ref{sec:mnist} evaluate PRoP's ability to keep the user's personalized policy private from others. However, one use case for PRoP is for \textit{distributed} personalization across users. Each user has a {unique} personalized behavior that they would like the robot to have that must not be accessible to other users.
In this experiment, we consider the influence of the size of the set $K^\star$ on performance, with each key corresponding to a separate personalized policy.
We use the experiments shown in \ref{sec:mnist} and increase the number of target policies and corresponding keys. We use the same key dimension and number of trainable parameters. Instead of the user's objective being a mapping to an arbitrary offset $k$ such that $l \mapsto (l + k) \mod 10$, each target user has a separate mapping from $l \mapsto l^\prime$ that they would like the robot to convey. We ensure that these mappings are bijective and i.i.d.. 
}

\subsection{Results}\label{sec:sim-results}

\begin{figure}
    %\vspace{-1.5em}
    \centering    
    \includegraphics[width=1.0\linewidth]{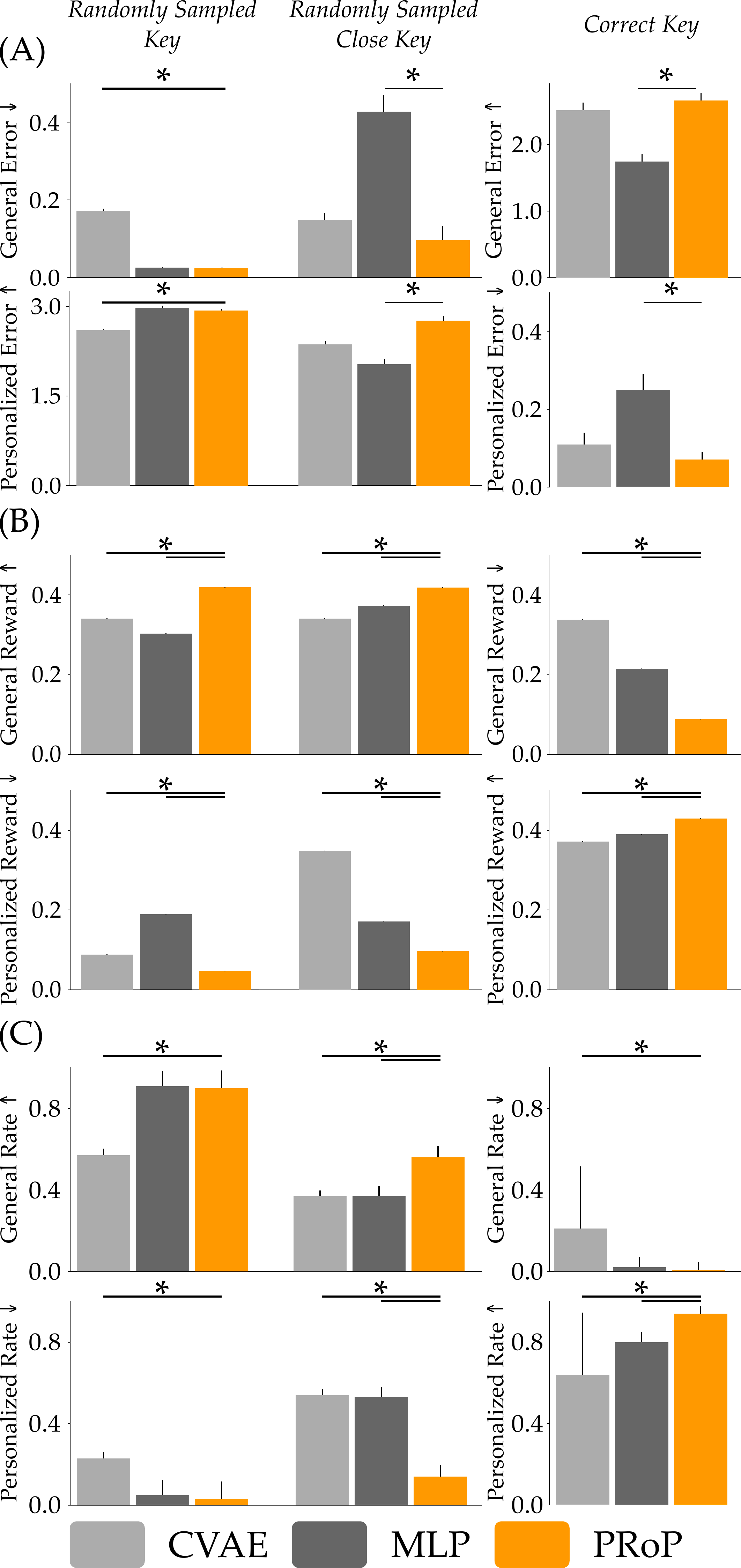}
    \caption{Results from our controlled simulations without pretrained policies. 
    For each experiment, the performance of the methods with respect to the general objective is shown in the first row and the performance with respect to the personalized objective is shown in the second row. 
    (A): In \textit{Imitation Learning}, the objective is to minimize mean-squared error between predicted and actual actions. 
    (B): In \textit{Reinforcement Learning}, the objective is to maximize normalized reward.
    (C): In \textit{Image Classification}, the objective is to maximize classification rate.
    Each column corresponds to a different key, with left corresponding to randomly sampled keys, center to the user's key with one bit flipped, and the final column to the user's key. 
    An asterisk indicates significance ($p < 0.05$) and an arrow indicates the desired trend.
    }
    \label{fig:sims1}
\end{figure}

Across all simulations and architectures, optimal performance is comprised of two elements: (a) alignment with the user's personalization when the key is \textit{correct}, and (b) alignment with $\pi^\star$ when the key is \textit{incorrect}. 
With this in mind, we present the results of our controlled simulations in \fig{sims1}. 
Results are averaged over $100$ simulated experiments and the vertical bars represent standard-error.
Each column of \fig{sims1} corresponds to a different key $k$. 
The first column corresponds to a randomly sampled key, i.e. $k \in \mathcal{K}_1$. The second column corresponds to a key that is one bit removed from the user's key, i.e. $k \in \mathcal{K}_2$. 
The final column corresponds to the user's key.
An optimal method would have high performance for the general objective in the first two columns and poor performance in the third column. Likewise, an optimal method would have high performance in the third column for the personalized objective and poor performance in the first two columns.

We find that PRoP has a statistically significant performance edge in all environments and keys when compared to baselines ($p < 0.05$). 
Notably, keys that are one-bit away from the correct key are less likely to leak user information than baselines. This distinction is important: if the rate of information leakage monotonically decreases with the number of bits incorrect in the key and the information leakage for close keys is low, then the model is difficult to attack. 
We find that in this critical point, our method exhibits far less information leakage than baselines. 
In simulations with relatively low dimensionality (e.g., \textit{Imitation Learning} (3-DoF)), PRoP generally performs on-par with a multi-layer perceptron model. However, as the dimensionality increases, the performance gap grows. In \textit{Reinforcement Learning}, the performance gap for correct keys between PRoP and MLP is substantial. {In \textit{Image Classification}, PRoP performs as well as MLP for randomly sampled keys, but PRoP is far better at personalizing to user data and maintaining their privacy. 
}

\begin{figure}[t]
    \centering
    \includegraphics[width=0.8\linewidth]{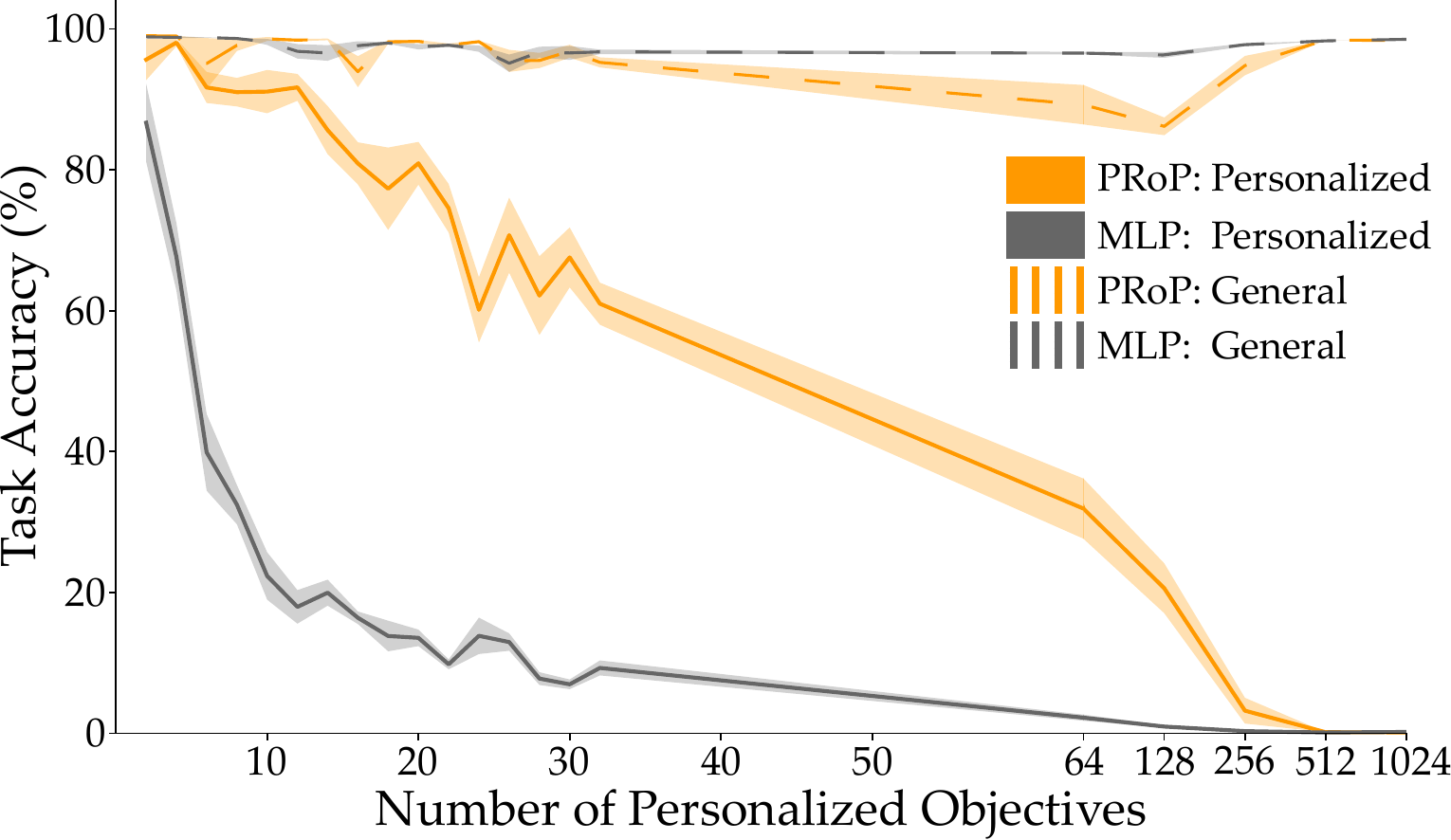}
    \caption{PRoP trained across multiple users with separate objectives achieves a higher personalization rate than alternatives. MLP outperforms in terms of \textit{Fallback} task performance, but is unable to personalize to specific user objectives. Note that beyond $64$ users, the $x$-axis scaling is logarithmic. The performance of PRoP is consistent until about $16$ users. After this threshold, PRoP's performance decays linearly until $512$ users. Alternatives such as an MLP or CVAE exhibit {exponential} performance decay with respect to the number of personalized objectives. These results indicate that PRoP can be used to compress a few users' personalizations into a {shared} network without leaking their information and without degrading overall task performance.}
    \label{fig:cap}
    \vspace{-1.0em}
\end{figure}

{\fig{cap} shows the results from Section~\ref{sec:cap}. Overall, we find that the performance of MLP and CVAE are nearly identical for $\lvert K^\star \rvert > 1$. PRoP safely personalizes to the users' objectives up to about $X$ users; after this point the model begins to fail. We stress that PRoP is personalizing using \textit{a single} key encoder across all users. With no additional memory overhead, PRoP is able to scale up to $X$ users with a single network. 
Implementation details and further analysis are available on our \href{https://prop-icra26.github.io}{project site}, such as simulations involving language personalization and weight obfuscation.
}

\section{User Study}\label{sec:us}

The controlled simulations in Section~\ref{sec:sims} suggest that our method successfully personalizes robot policies to specific user behaviors in a private manner.
It surpasses baseline performance while conforming to the architecture of the pretrained network. 
To evaluate how {PRoP} performs with real users, we next conducted an in-person user study with $N = 12$ participants.
This task was a mock kitchen environment. 
The robot was tasked with assembling different meals, and the order of ingredients was personalized to different users according to their password. 

\p{Experimental Setup}
Participants operated in a mock kitchen environment alongside a Universal Robotics UR-10 robotic manipulator with Robotiq gripper. 
Each participant was tasked with logging into a website using their chosen username and password, then ordering a personal sandwich order from a collection of ingredients. 
Upon receiving their order, three policies were trained: CVAE, MLP, and PRoP. 
Each policy then assembled sandwiches in the real-world with bit-represented keys: the user's key, a randomly sampled key, and a key that differed from the user's by one bit. 

\p{Participants and Procedure}
We recruited $12$ participants of our university community (average age $23 \pm 3.4$). Of the $12$ participants, $3$ did not have experience with robotics and $8$ did not have experience with robot learning.
Participants received monetary compensation for their time and provided informed written consent (IRB \#20-755). 

We leveraged a between-subjects design where every participant provided an order and observed the interaction between their key and two randomly sampled alternative keys from previous users. 
The user's goal was to correctly predict which order corresponded to their key. Participants were never told which order corresponded to their key nor what method was in operation. 

\p{Dependent Measures}
We recorded the key, sandwich order, states, and actions at each timestep during the interaction.
To assess performance, we consider an interaction correct if the order corresponds to the key's order in the unified dataset $\{(k_\text{user}, o_\text{user}\} \cup \{(k_1, o_1), \ldots, (k_N, o_N)\}$ where $k$ is the key and $o$ is the order.
Before training we ensure that $o_\text{user}$ and $k_\text{user}$ do not exist in the default dataset. 
We use the following metric as a proxy to performance: 
\begin{equation}
\text{Score}  
=
\quad
\begin{gathered}
\mathbb{I}\left(k = k_\text{user} ~\land~ o=o_\text{user} \right)
\\
+ 
\mathbb{I}\left(k \ne k_\text{user} ~\land~ o = o_k \right)
\end{gathered}
\label{eq:us-score}
\end{equation}
where $\mathbb{I}$ is the indicator function.
The \textit{Score} metric is higher when the network provides the correct order for the corresponding key and is lower when the network fails to personalize to the user's key. 
An optimal value for this metric is $3$ (since we roll-out each method for $3$ different keys).
We use a similar metric as a proxy to information leakage:
\begin{equation}
\text{Privacy}
=
\quad
\begin{gathered}
\mathbb{I}\left(k = k_\text{user} ~\land~ o \ne o_\text{user} \right)
\\
+ 
\mathbb{I}\left(k \ne k_\text{user} ~\land~ o = o_\text{user} \right)
\end{gathered}
\label{eq:us-priv}
\end{equation}
Analogous to \eq{us-score}, this metric is maximized when the user's order is leaked: for example, if the order is present for multiple user keys. 
Likewise, if the user receives an order for their key that is not their intended order, this metric increases. 
An optimal value for this metric is $0$.

\p{Hypothesis}
We had two hypotheses for this user study:
\begin{quote}
    \textbf{H1.} \quad \textit{PRoP will} \textbf{personalize} \textit{more effectively than baselines.}
    \\
    \textbf{H2.} \quad \textit{The personalization that PRoP provides will be more} \textbf{private} \textit{than baselines.}
\end{quote}

\begin{figure}[t]
    \centering
    \includegraphics[width=1.0\linewidth]{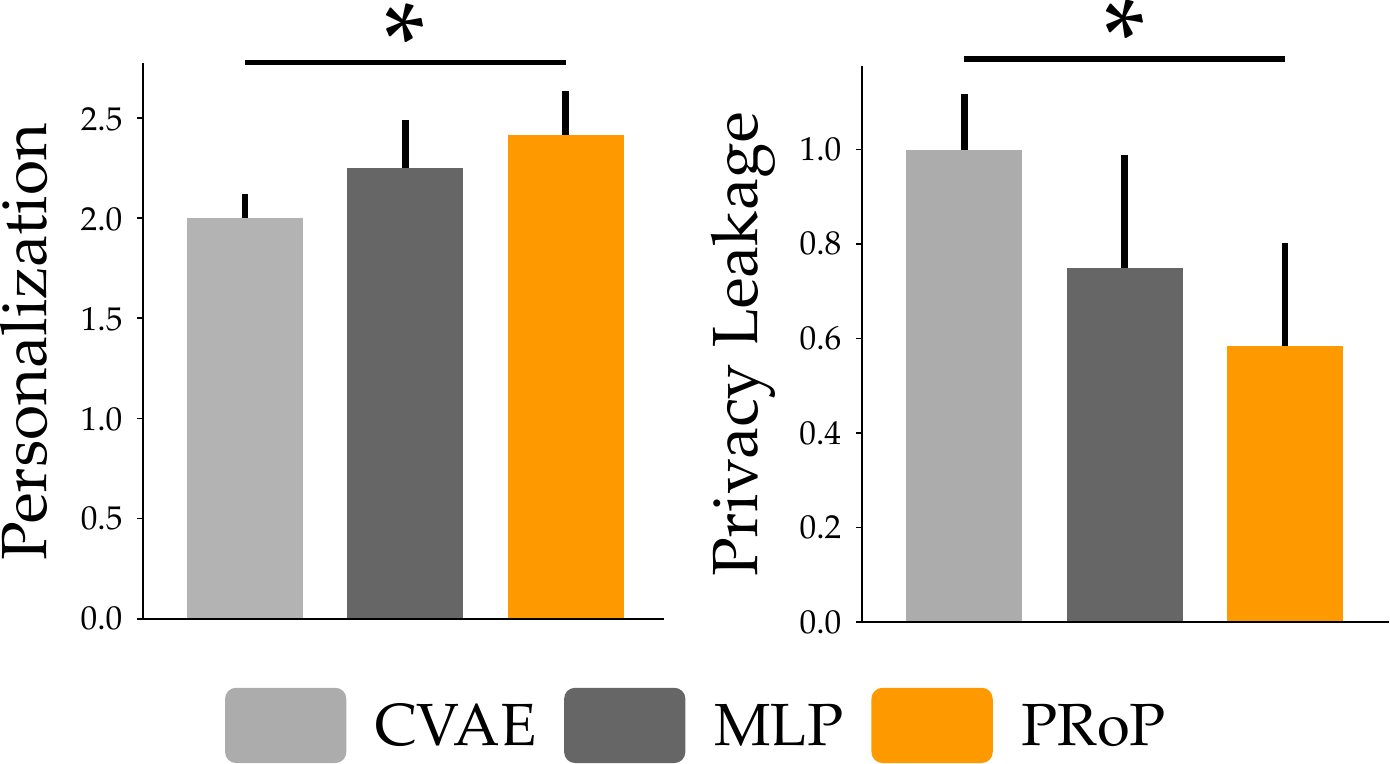}
    \caption{Results from our in-person user study. PRoP outperforms baselines in terms of average privacy leakage and personalization. 
    Note that \textit{personalization} should be high, while \textit{privacy leakage} should be low. 
    Error bars show standard error and an asterisk (*) indicates significance ($p < 0.05$).
    (Left) PRoP enables better personalization than baselines while using a similar number of learnable parameters. This metric is evaluated using \eq{us-score}.
    (Right) Likewise, PRoP is less prone to information leakage than baselines. This metric is evaluated using \eq{us-priv}.
    }
    \label{fig:us}
    \vspace{-1.0em}
\end{figure}

\p{Results}
We used a key size of $32$ for this user study.
The results from our in-person user study are summarized in \fig{us}. 
To assess \textbf{H1}, we consult the Score metric from \eq{us-score}. We find that PRoP outperforms baselines in terms of average \textit{Score} across interactions, although this performance improvement is not statistically significant across all methods \fig{us} (Right). 
Likewise, for \textbf{H2}, we consult the \textit{Privacy} metric from \eq{us-priv}.
From \fig{us} (Left), we can see that PRoP leaks less user information to alternative keys, and is significantly lower than CVAE ($p < 0.05$). 
%but the average information leakage between PRoP and MLP is not statistically significant. As in Section~\ref{sec:sims}, minimizing the information leakage for keys that are one bit removed from the user's key is incredibly important. We find that with PRoP, this leakage is minimal and monotonically decreases as the number of incorrect bits increases. 
For further analysis, consult our \href{https://prop-icra26.github.io/}{project site}.

%\begin{figure}[t]
%    \centering
%    \includegraphics[width=1.0\linewidth]{figures/us.pdf}
%    \caption{Results from our in-person user study. PRoP outperforms baselines in terms of average privacy leakage and personalization. 
%    Note that \textit{personalization} should be high, while \textit{privacy leakage} should be low. 
%    Error bars show standard error and an asterisk (*) indicates significance ($p < 0.05$).
%    (Left) PRoP enables better personalization than baselines while using a similar number of learnable parameters. This metric is evaluated using \eq{us-score}.
%    (Right) Likewise, PRoP is less prone to information leakage than baselines. This metric is evaluated using \eq{us-priv}.
%    }
%    \label{fig:us}
%    \vspace{-1.0em}
%\end{figure}
%
\section{Conclusion}\label{sec:conclusion}

In this manuscript we present {PRoP}: a method to personalize pretrained neural network policies for new users. 
{PRoP} preserves the architecture and behavior of the original network while enabling personalization to specific users in a private fashion. 
Our method can also be used in an end-to-end manner --- without any pretrained model. 
In imitation learning, reinforcement learning, classification, and in real-world user studies, {PRoP} outperforms baselines in terms of \textit{personalization} and \textit{privacy}.
Prior HRI literature achieves personalization, but PRoP is unique in that it provides personalization, preserves pretrained models, and prevents other agents from learning the user's preferences. 
Through our simulated and real-world experiments we find that PRoP performs well across various applications such as imitation learning, reinforcement learning, image classification. 
%{PRoP}'s mechanism for intermediate augmentation also shows promise in large-language models and for weight obfuscation; see Appendixes~\ref{app:lang} and \ref{app:encrypt} for details.

%%%%%%%%%%%%%%%%%%%%%%%%%%%%%%%%%%%%%%%%%%%%%%%%%%%%%%%%%%%%%%%%%%%%%%%%%%%%%%%%%

\bibliographystyle{IEEEtran}
\bibliography{bibtex}

@article{yang2024does,
  title={Does negative sampling matter? a review with insights into its theory and applications},
  author={Yang, Zhen and Ding, Ming and Huang, Tinglin and Cen, Yukuo and Song, Junshuai and Xu, Bin and Dong, Yuxiao and Tang, Jie},
  journal={IEEE Transactions on Pattern Analysis and Machine Intelligence},
  volume={46},
  number={8},
  pages={5692--5711},
  year={2024},
  publisher={IEEE}
}

@misc{news,
author={Erin Relford},
howpublished={\url{https://iapp.org/news/a/privacy-in-the-age-of-robotics?}},
year={2025}
}

@article{jevtic2018personalized,
  title={Personalized robot assistant for support in dressing},
  author={Jevti{\'c}, Aleksandar and Valle, Andr{\'e}s Flores and Aleny{\`a}, Guillem and Chance, Greg and Caleb-Solly, Praminda and Dogramadzi, Sanja and Torras, Carme},
  journal={IEEE Trans. Cognitive and Developmental Sys.},
  year={2018},
}

@article{losey2022physical,
  title={Physical interaction as communication: {L}earning robot objectives online from human corrections},
  author={Losey, Dylan P and Bajcsy, Andrea and O’Malley, Marcia K and Dragan, Anca D},
  journal={IJRR},
  year={2022},
}

@article{kawaharazuka2024real,
  title={Real-world robot applications of foundation models: A review},
  author={Kawaharazuka, Kento and Matsushima, Tatsuya and Gambardella, Andrew and Guo, Jiaxian and Paxton, Chris and Zeng, Andy},
  journal={Advanced Robotics},
  year={2024},
}

@inproceedings{sohn2015learning,
  title={Learning Structured Output Representation using Deep Conditional Generative Models},
  author={Sohn, Kihyuk and Lee, Honglak and Yan, Xinchen},
  booktitle={NeurIPS},
  year={2015}
}

@article{schulman2017proximal,
  title={Proximal policy optimization algorithms},
  author={Schulman, John and Wolski, Filip and Dhariwal, Prafulla and Radford, Alec and Klimov, Oleg},
  journal={arXiv preprint arXiv:1707.06347},
  year={2017}
}

@article{deng2012mnist,
  title={The {MNIST} database of handwritten digit images for machine learning research},
  author={Deng, Li},
  journal={IEEE Signal Processing Magazine},
  year={2012},
}

@article{gallouedec2021pandagym,
  title        = {panda-gym: {O}pen-Source Goal-Conditioned Environments for Robotic Learning},
  author       = {Gallou{\'e}dec, Quentin and Cazin, Nicolas and Dellandr{\'e}a, Emmanuel and Chen, Liming},
  year         = 2021,
  journal      = {NeurIPS},
}

@article{black2024pi_0,
  title={$\pi_0$: A Vision-Language-Action Flow Model for General Robot Control},
  author={Black, Kevin and Brown, Noah and Driess, Danny and Esmail, Adnan and Equi, Michael and Finn, Chelsea and Fusai, Niccolo and Groom, Lachy and Hausman, Karol and Ichter, Brian and others},
  journal={arXiv preprint arXiv:2410.24164},
  year={2024}
}

@article{li2024cogact,
  title={Cog{ACT}: {A} foundational vision-language-action model for synergizing cognition and action in robotic manipulation},
  author={Li, Qixiu and Liang, Yaobo and Wang, Zeyu and Luo, Lin and Chen, Xi and Liao, Mozheng and Wei, Fangyun and Deng, Yu and Xu, Sicheng and Zhang, Yizhong and others},
  journal={arXiv preprint arXiv:2411.19650},
  year={2024}
}

@inproceedings{lee2011understanding,
  title={Understanding users' perception of privacy in human-robot interaction},
  author={Lee, Min Kyung and Tang, Karen P and Forlizzi, Jodi and Kiesler, Sara},
  booktitle={ACM/IEEE International Conference on Human-Robot Interaction},
  year={2011}
}

@article{chatzimichali2020toward,
  title={Toward privacy-sensitive human--robot interaction: Privacy terms and human--data interaction in the personal robot era},
  author={Chatzimichali, Anna and Harrison, Ross and Chrysostomou, Dimitrios},
  journal={Journal of Behavioral Robotics},
  year={2020},
}

@article{jain2015learning,
  title={Learning preferences for manipulation tasks from online coactive feedback},
  author={Jain, Ashesh and Sharma, Shikhar and Joachims, Thorsten and Saxena, Ashutosh},
  journal={The International Journal of Robotics Research},
  year={2015},
}

@article{gasteiger2023factors,
  title={Factors for personalization and localization to optimize human--robot interaction: A literature review},
  author={Gasteiger, Norina and Hellou, Mehdi and Ahn, Ho Seok},
  journal={International Journal of Social Robotics},
  year={2023},
}

@article{mehta2024unified,
  title={Unified Learning from Demonstrations, Corrections, and Preferences during Physical Human--Robot Interaction},
  author={Mehta, Shaunak A and Losey, Dylan P},
  journal={ACM Transactions on Human-Robot Interaction},
  year={2024},
}

@article{christie2024limit,
  title={{LIMIT: L}earning interfaces to maximize information transfer},
  author={Christie, Benjamin A and Losey, Dylan P},
  journal={ACM Transactions on Human-Robot Interaction},
  year={2024},
}

@article{hu2023llm,
  title={Llm-adapters: An adapter family for parameter-efficient fine-tuning of large language models},
  author={Hu, Zhiqiang and Wang, Lei and Lan, Yihuai and Xu, Wanyu and Lim, Ee-Peng and Bing, Lidong and Xu, Xing and Poria, Soujanya and Lee, Roy Ka-Wei},
  journal={arXiv preprint arXiv:2304.01933},
  year={2023}
}

@inproceedings{yang2024robot,
  title={Robot fine-tuning made easy: {P}re-training rewards and policies for autonomous real-world reinforcement learning},
  author={Yang, Jingyun and Mark, Max Sobol and Vu, Brandon and Sharma, Archit and Bohg, Jeannette and Finn, Chelsea},
  booktitle={ICRA},
  year={2024},
}

@inproceedings{triastcyn2020bayesian,
  title={Bayesian differential privacy for machine learning},
  author={Triastcyn, Aleksei and Faltings, Boi},
  booktitle={ICML},
  year={2020},
}

@inproceedings{lytvyn2019information,
  title={Information encryption based on the synthesis of a neural network and AES algorithm},
  author={Lytvyn, Vasyl and Peleshchak, Ivan and Peleshchak, Roman and Vysotska, Victoria},
  booktitle={AICT},
  year={2019},
}

@article{lee2022privacy,
  title={Privacy-preserving machine learning with fully homomorphic encryption for deep neural network},
  author={Lee, Joon-Woo and Kang, HyungChul and Lee, Yongwoo and Choi, Woosuk and Eom, Jieun and others},
  journal={IEEE Access},
  year={2022},
}

@article{pulido2021privacy,
  title={Privacy-preserving neural networks with homomorphic encryption: Challenges and opportunities},
  author={Pulido-Gaytan, Bernardo and Tchernykh, Andrei and Cort{\'e}s-Mendoza, Jorge M and Babenko, Mikhail and Radchenko, Gleb and Avetisyan, Arutyun and Drozdov, Alexander Yu},
  journal={Peer-to-Peer Networking and Applications},
  year={2021},
}

@inproceedings{zhang2021leakage,
  title={Leakage of dataset properties in $\{$Multi-Party$\}$ machine learning},
  author={Zhang, Wanrong and Tople, Shruti and Ohrimenko, Olga},
  booktitle={USENIX Security},
  year={2021}
}

@article{chen2020comprehensive,
  title={A comprehensive analysis of information leakage in deep transfer learning},
  author={Chen, Cen and Wu, Bingzhe and Qiu, Minghui and Wang, Li and Zhou, Jun},
  journal={arXiv preprint arXiv:2009.01989},
  year={2020}
}

@inproceedings{abadi2016deep,
  title={Deep learning with differential privacy},
  author={Abadi, Martin and Chu, Andy and Goodfellow, Ian and McMahan, H Brendan and Mironov, Ilya and Talwar, Kunal and Zhang, Li},
  booktitle={ACM Conference on Computer and Communications Security},
  year={2016}
}

@article{li2019differentially,
  title={Differentially private meta-learning},
  author={Li, Jeffrey and Khodak, Mikhail and Caldas, Sebastian and Talwalkar, Ameet},
  journal={arXiv preprint arXiv:1909.05830},
  year={2019}
}

@InProceedings{pmlr-v164-hundt22a,
  title = 	 {"Good Robot! Now Watch This!": Repurposing Reinforcement Learning for Task-to-Task Transfer},
  author =       {Hundt, Andrew and Murali, Aditya and Hubli, Priyanka and Liu, Ran and Gopalan, Nakul and Gombolay, Matthew and Hager, Gregory D.},
  booktitle = 	 {Proceedings of the 5th Conference on Robot Learning},
  pages = 	 {1564--1574},
  year = 	 {2022},
  volume = 	 {164},
}

@inproceedings{kim2024robust,
  title={Robust policy learning via offline skill diffusion},
  author={Kim, Woo Kyung and Yoo, Minjong and Woo, Honguk},
  booktitle={Proceedings of the AAAI Conference on Artificial Intelligence},
  volume={38},
  number={12},
  pages={13177--13184},
  year={2024}
}

@INPROCEEDINGS{10610421,
  author={Yang, Jingyun and Mark, Max Sobol and Vu, Brandon and Sharma, Archit and Bohg, Jeannette and Finn, Chelsea},
  booktitle={2024 IEEE International Conference on Robotics and Automation (ICRA)}, 
  title={Robot Fine-Tuning Made Easy: Pre-Training Rewards and Policies for Autonomous Real-World Reinforcement Learning}, 
  year={2024},
  volume={},
  number={},
  pages={4804-4811},
  doi={10.1109/ICRA57147.2024.10610421}}

@article{li2024process,
  title={Process reward model with q-value rankings},
  author={Li, Wendi and Li, Yixuan},
  journal={arXiv preprint arXiv:2410.11287},
  year={2024}
}

@inproceedings{yu2024b,
  title={B-coder: Value-based deep reinforcement learning for program synthesis},
  author={Yu, Zishun and Tao, Yunzhe and Chen, Liyu and Sun, Tao and Yang, Hongxia},
  booktitle={12th International Conference on Learning Representations, ICLR 2024},
  year={2024}
}

@article{julian2020never,
  title={Never stop learning: The effectiveness of fine-tuning in robotic reinforcement learning},
  author={Julian, Ryan and Swanson, Benjamin and Sukhatme, Gaurav S and Levine, Sergey and Finn, Chelsea and Hausman, Karol},
  journal={arXiv preprint arXiv:2004.10190},
  year={2020}
}

@article{hubotter2024active,
  title={Active few-shot fine-tuning},
  author={H{\"u}botter, Jonas and Sukhija, Bhavya and Treven, Lenart and As, Yarden and Krause, Andreas},
  journal={arXiv preprint arXiv:2402.15441},
  year={2024}
}

@article{bagatella2024active,
  title={Active fine-tuning of multi-task policies},
  author={Bagatella, Marco and H{\"u}botter, Jonas and Martius, Georg and Krause, Andreas},
  journal={arXiv preprint arXiv:2410.05026},
  year={2024}
}

\end{document}